# Style Transfer of Black and White Silhouette Images using CycleGAN and a Randomly Generated Dataset

Worasait Suwannik
*Department of Computer Science*
*Kasetsart University*
Bangkok, Thailand
worasait.suwannik@gmail.com

*Abstract*— CycleGAN can be used to transfer an artistic style to an image. It does not require pairs of source and stylized images to train a model. Taking this advantage, we propose using randomly generated data to train a machine learning model that can transfer traditional art style to a black and white silhouette image. The result is noticeably better than the previous neural style transfer methods. However, there are some areas for improvement, such as removing artifacts and spikes from the transformed image.

## I. Introduction

Modern machine learning models produce awesome results. Nowadays, it is not difficult to build a machine learning model. The machine learning source code is easily obtained from a blog, a tutorial, an open-source repository such as GitHub or shared through Google Colab. Moreover, cloud services provide powerful GPU runtime to build the machine learning model. However, many researchers and organizations face the problem of lacking training data, especially in a particular domain or with no platform to collect data generated by humans.

One domain that lacks an open dataset is traditional Thai art. Thai traditional decorative pattern or *lai thai* (ลายไทย) appears in mural paintings, gold leaf doors, engraved furniture, and many others. Lai thai is inspired by nature such as flowers, leaves, vines, and flame. One important example of lai thai is the Kanok pattern (ลายกนก) as shown in Figure 1.

We aim to use artificial intelligence to transform a black and white image of a single prominent object into a traditional Thai decorative pattern (see Figure 2). Training a machine learning model for the task requires a dataset with pairs of a source image and a stylized image. However, we cannot find a dataset that matches this requirement. Therefore, we propose using randomly generated data to train a machine learning model.

## II. Related Work

### A. Neural Style Transfer

Artificial neural networks can be used to transfer a style of a reference image to another image known as a content image [1]. They are often used to transfer styles of famous artists or painting genres such as Ukiyo-e, Vincent van Gogh, or Monet. The results are astonishing. The photo is turned into a painting of the famous style.

We tried to transfer the style of our randomly generated image to the silhouette image of a bird using the code from TensorFlow tutorial. Figure 3 (a) (b) show the output from TensorFlow hub model using Fig 1 and its inverted version (white background) as style images. Figure 3 (c) (d) show the final output from the tutorial code. As seen from the output, these machine learning models are unsuitable for this task.

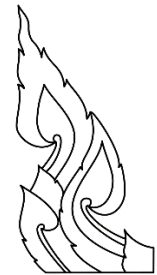

Fig. 1. Kanok pattern. This image is slightly modified from a public domain image obtained from Wikipedia.

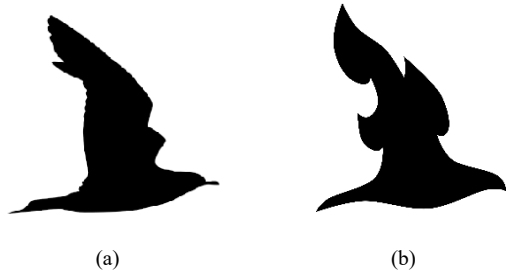

Fig. 2. Transform an image into a Thai traditional art style. (a) an original image (b) target image (manually drawn).

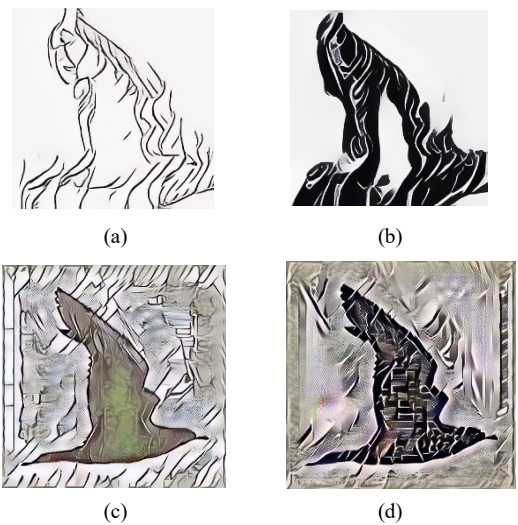

Fig. 3. Using neural style transfer in TensorFlow tutorial to stylize an image from Fig 2.a using a style from Fig 1 and its inverted version.

## B. Pix2Pix

Generative Adversarial Network or GAN [2] can be used to generate images. GAN consists of a generator network and a discriminator network. The generator and the discriminator compete against each other and use each other to improve their performance. The discriminator predicts whether the input is real or fake (i.e., generated from the discriminator.) The generator tries to fool the discriminator (i.e., to make the discriminator predict that the generated output is real).

Pix2pix [3] is a GAN that can perform an image-to-image translation task. To train the model, pairs of a source image and a target image are needed. Some dataset is not difficult to create. For example, in a colorize a black and white image task, its dataset is a pair of a color image and its desaturated counterpart.

Google provides TensorFlow pix2pix code via their Colab service. The architecture of a generator and a discriminator are U-net and a convolutional PatchGAN classifier. U-net consists of encoder and a decoder. The encoder consists of downsampling layers. The decoder consists of upsampling layers. There are skip connections between the encoder and the decoder.

For our task, we cannot use Pix2Pix because we do not have a dataset containing pairs of a source image and a stylized image.

## C. CycleGAN

CycleGAN [4] can also perform image-to-image translation. However, it has an advantage over Pix2Pix in that CycleGAN does not require a pair of a source and a target image to train the model. In CycleGAN, there are 2 generators and 1 discriminator.

Google also provides TensorFlow CycleGAN code via their Colab service. In the code, the loss function related to generators consists of three terms: identity loss, cycle loss, and generator loss. Before we explain those losses, let's define each component in CycleGAN.

$X_A$ is a real image in domain A

$G_A$ is a generator that generates a fake image in domain A. It normally receives input from domain B.

$X_B$ and $G_B$ are defined analogously.

The cycle loss is calculated by feeding input $X_A$ to the generator $G_B$ and then feed the output (a fake image of domain B) to the generator $G_A$. In other words, the image is transformed to another domain and transformed back to the original domain. After completing the cycle, the final output should be the same as the original input because both generators only change the style of images. The pixel difference between the final output and the original input is the cycle loss.

The generator loss is binary cross entropy loss (BCE loss) between the discriminator output and 1s.

The identity loss is the loss measured when feeding an input $X_A$ to the generator $G_A$ that generates an image of the same domain and measures the distance between the input and the output. The rationale is that the generator $G_A$ should an output image in domain A whether the input is in domain A or domain B. Note that the generator $G_A$ receives input $X_B$, which is an image in another domain. For color images, this loss term helps preserve color.

## III. EXPERIMENT

### A. Domain A: Bird Images

There are two image domains in our work. The first domain is black and white images with one prominent object. The second domain contains the black and white Thai traditional images. For the first domain, we obtain silhouette images from phylopic.org using its API. We select only bird images to make it easier to train the generators. The training and testing set have 300 and 50 images, respectively.

### B. Domain B: Style Images

For the black and white Thai traditional image domain, we cannot find image data set of Thai traditional patterns that matches our requirements. Therefore, we randomly generated images.

First, we select some components of Kanok pattern (see Figure 1) and add variations to those components. The result is 32 basic elements (see Figure 4).

After that, we randomly combine those elements. The selected elements are randomly resized, mirrored, and rotated. Since images in the first domain have margins, the selected elements are randomly placed inside the margin. Sample results are shown in Figure 5 The training and testing set have 2,900 and 100 images, respectively.

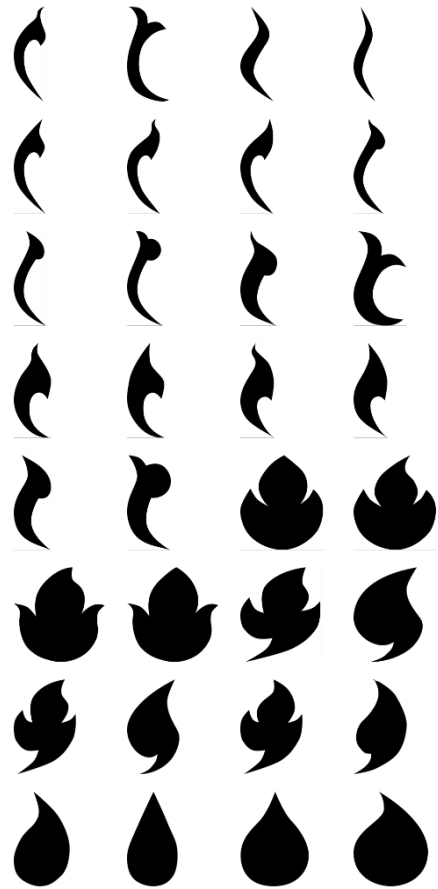

Fig. 4. Basic elements.

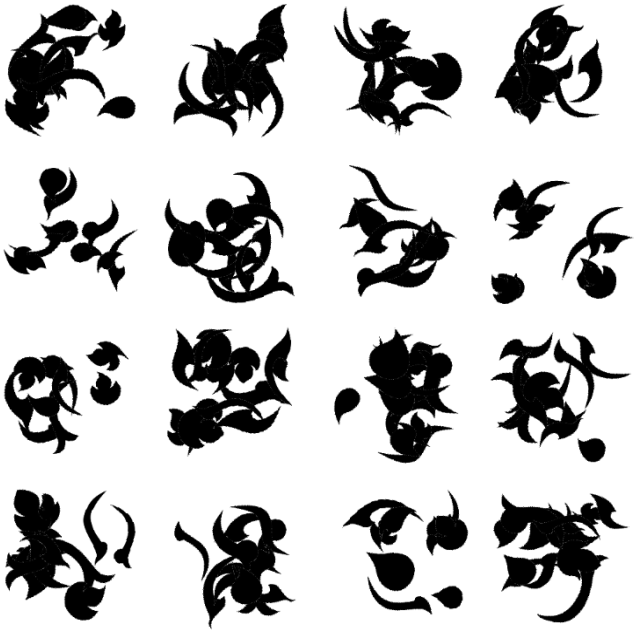

Fig. 5. Sample style images.

*C. Architecture*

We use the CycleGAN code from TensorFlow tutorial. This implementation is different from the CycleGAN paper in the network architecture. The tutorial uses a modified U-net, while the CycleGAN paper use a modified ResNet. In our experiment, the training data is not randomly jittered (i.e., resize, randomly cropping, and random mirroring). We train the model for 200 epochs.

## IV. Result

Figure 8 shows the output of the generators obtained from the first 10 testing data. The first column contains images obtained from phylopic.org. The second column, labeled baseline, contains output from CycleGAN. Other columns contain output obtained from changing weights of various losses. For example, 2xCycleLos means the weight of cycle loss is 2.

The result is not what we expected when we started this project. First, artifacts appear in many images. Second, the object contains white parts. However, CycleGAN can dutifully transfer the style of the randomly generated images to the bird image better than the previous neural style transfer mentioned in this paper (see Figure 6 and 7).

To improve the result in the future, we might use CycleGAN to automatically create an image-pair dataset by sending the output of CycleGAN to the following process: removing white parts in the object and removing artifacts. After that, the dataset would be used to train a pix2pix model.

## V. Conclusion

This paper proposed the use of randomly generated images to train a CycleGAN model. The task is to stylize a black and white image. To make the transfer closer to our expectation, weights of loss function should be adjusted. There are areas for improvement such as removing artifacts and spikes in a transformed image.

### Acknowledgment

We would like to thank the authors of silhouette images of birds and phylopic.org for providing API for uploading the images.

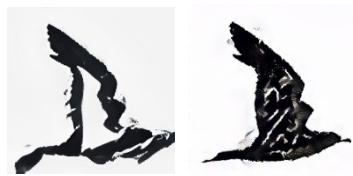

Fig. 6. Output from neural style transfer.

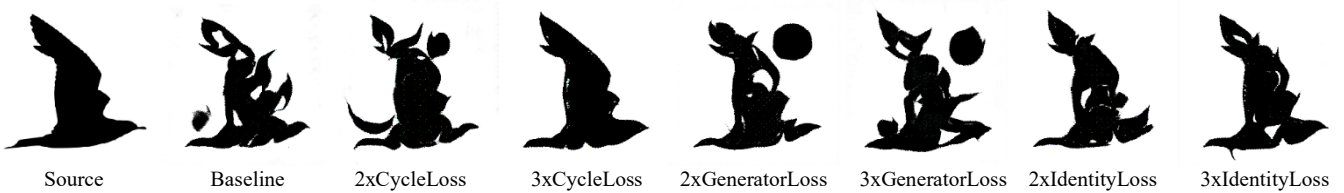

Source    Baseline    2xCycleLoss    3xCycleLoss    2xGeneratorLoss    3xGeneratorLoss    2xIdentityLoss    3xIdentityLoss

Fig. 7. Output from CycleGAN

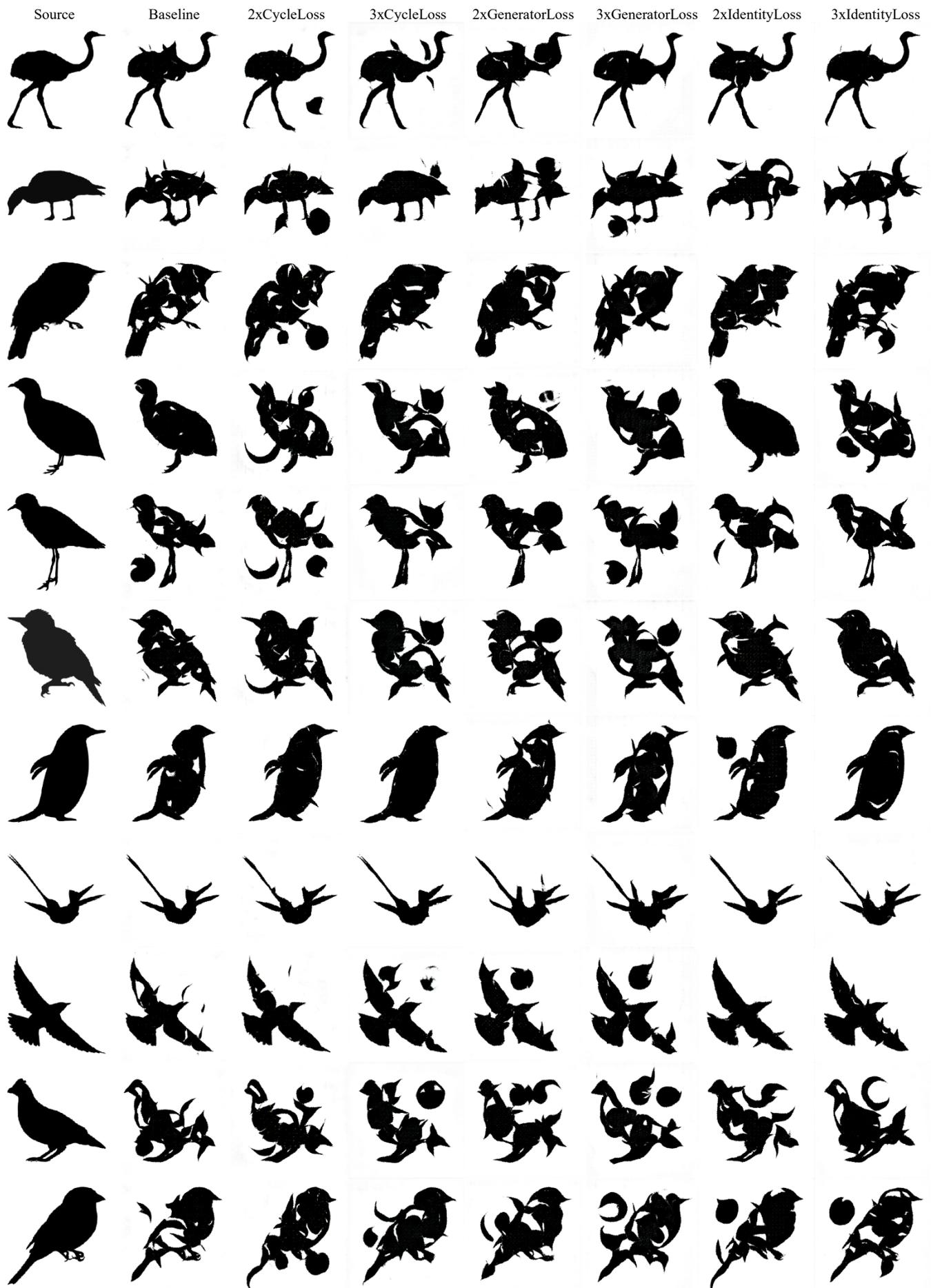

Fig. 8. Comparing various weights and losses.